\def\year{2022}\relax
\documentclass[letterpaper]{article} 
\usepackage{aaai22}  
\usepackage{times}  
\usepackage{helvet}  
\usepackage{courier}  
\usepackage[hyphens]{url}  
\usepackage{graphicx} 
\usepackage{natbib}  
\usepackage{caption} 
\DeclareCaptionStyle{ruled}{labelfont=normalfont,labelsep=colon,strut=off} 
\usepackage{algorithm}
\usepackage{algorithmic}

\usepackage{newfloat}
\usepackage{listings}
\floatstyle{ruled}
\newfloat{listing}{tb}{lst}{}
\floatname{listing}{Listing}

\usepackage{amsmath}
\usepackage{amssymb}
\usepackage{multirow}

\title{Procedural Text Understanding via Scene-Wise Evolution}
\author {
    Jialong Tang\textsuperscript{\rm 1,3},
    Hongyu Lin\textsuperscript{\rm 1},
    Meng Liao\textsuperscript{\rm 4,*},
    Yaojie Lu\textsuperscript{\rm 1,3},\\
    Xianpei Han\textsuperscript{\rm 1,2},
    Le Sun\textsuperscript{\rm 1,2,*},
    Weijian Xie\textsuperscript{\rm 4},
    Jin Xu\textsuperscript{\rm 4}
}
\affiliations {
    \textsuperscript{\rm 1} Chinese Information Processing Laboratory, Beijing, China \\ 
    \textsuperscript{\rm 2} State Key Laboratory of Computer Science Institute of Software, Chinese Academy of Sciences, Beijing, China\\
    \textsuperscript{\rm 3} University of Chinese Academy of Sciences, Beijing, China\\
    \textsuperscript{\rm 4} Data Quality Team, WeChat, Tencent Inc., China\\
    \{jialong2019,hongyu,yaojie2017,xianpei,sunle\}@iscas.ac.cn \\ \{maricoliao, vikoxie, jinxxu\}@tencent.com
}

\usepackage{bibentry}

\begin{document}

\maketitle

\begin{abstract}
{
  \renewcommand{\thefootnote}{\fnsymbol{footnote}}
}
Procedural text understanding requires machines to reason about entity states within the dynamical narratives.
Current procedural text understanding approaches are commonly \textbf{entity-wise}, which separately track each entity and independently predict different states of each entity. 
Such an entity-wise paradigm does not consider the interaction between entities and their states.
In this paper, we propose a new \textbf{scene-wise} paradigm for procedural text understanding, which jointly tracks states of all entities in a scene-by-scene manner.
Based on this paradigm, we propose \textbf{S}cene \textbf{G}raph \textbf{R}easoner (\textbf{SGR}), which introduces a series of dynamically evolving scene graphs to jointly formulate the evolution of entities, states and their associations throughout the narrative.
In this way, the deep interactions between all entities and states can be jointly captured and simultaneously derived from scene graphs.
Experiments show that SGR not only achieves the new state-of-the-art performance but also significantly accelerates the speed of reasoning. 
\let\thefootnote\relax\footnotetext{${}^{*}$Corresponding authors.}
\end{abstract}

\section{Introduction}

\begin{figure}[!t]
\centering
\includegraphics[width=0.95\columnwidth]{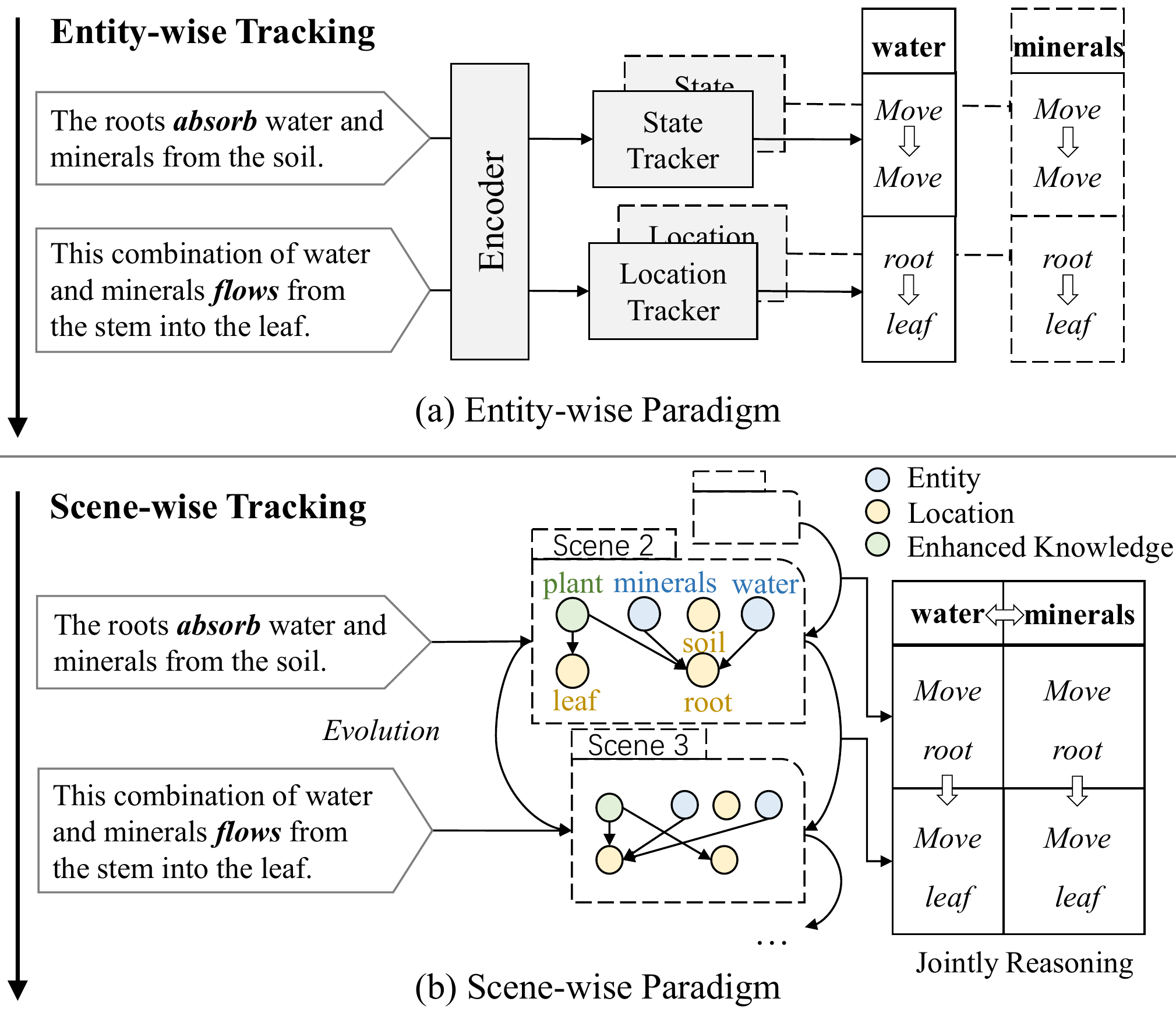}
\caption{Comparison between the traditional entity-wise paradigm and the proposed scene-wise paradigm for procedural text understanding.
We can see that: 
(a) the entity-wise paradigm tracks each entity separately, and predict state changes and locations of each entity independently;
(b) the scene-wise paradigm jointly tracks the state changes and locations of all entities scene-by-scene.}
\label{fig:introduction}
\end{figure}

Understanding how events will affect the world is the essence of intelligence~\cite{Henaff-2017-Tracking}.
Procedural text understanding, aiming to track the state changes (e.g., create, move, destroy) and locations (a span in the text) of entities throughout the whole procedure, is a representative task to estimate the machine intelligence on such ability~\cite{Mishra-2018-Tracking}. 
For example, in Figure~\ref{fig:introduction} (a), given a narrative describing the procedure of photosynthesis, as well as a pre-specified entity ``\emph{water}'', a procedural text understanding model is asked to predict the corresponding \{\emph{State}, \emph{location}\} sequences: \{\emph{Move}, \emph{root}\}, \{\emph{Move}, \emph{leaf}\}.
Compared with conventional factoid-style reading comprehension tasks~\cite{Seo-2017-Bidirectional,Clark-2018-Simple}, procedural text understanding is more challenging because it requires to model and reason with the dynamical world~\cite{Mishra-2018-Tracking,Bosselut-2018-Simulating}. 

Most approaches resolve procedural text understanding task in an \textbf{entity-wise} paradigm, where   
each entity is tracked separately, and state changes and locations of each entity are independently predicted.
Along this line, as Figure~\ref{fig:introduction} (a) shows, current procedural text understanding models mainly resort to hierarchical neural network architectures, which first encode the document-entity pair using a token-level encoder, then track the state changes and locations by two separate sentence-level trackers~\cite{Mishra-2018-Tracking,Mishra-2019-Everything,Du-2019-Consistent,Tang-2020-Understanding,Gupta-2019-Tracking}. 
More recently, the main research hot spot in this direction is how to obtain more effective document-entity representations by introducing graph-based architectures~\cite{Das-2019-Building,Zhong-2020-Heterogeneous,Huang-2021-Reasoning}, pretrained language models~\cite{Gupta-2019-Effective,Amini-2020-Procedural,Zhang-2020-Knowledge} or external knowledge bases~\cite{Ribeiro-2019-Predicting,Tandon-2018-Reasoning}.

Unfortunately, the traditional entity-wise paradigm ignores the interactions between different entities in the same narrative, as well as the associations between the state changes and locations of one entity. 
Specifically, the multiple entities mentioned in the same narrative are highly correlated with each other.
For example, if we know ``\emph{water}'' and ``\emph{minerals}'' will be combined into a ``\emph{mixture}'', we can confirm that they must in the same location ``\emph{leaf}''. 
Furthermore, the states and locations of an entity are highly associated. 
For example, if we know the location of ``\emph{water}'' changes from ``\emph{root}'' to ``\emph{leaf}'', we can easily predict the state of ``\emph{water}'' is ``\emph{Move}''.
Besides, the states/locations at current step depend on the states/location at previous steps. 
For example, if we know ``\emph{root}'' and ``\emph{leaf}'' are parts of ``\emph{plant}'', we will tend to predict the location ``\emph{leaf}'' after the location ``\emph{root}''.
However, current entity-wise paradigm is unable to exploit the above-mentioned interactions and associations.
In addition, reasoning procedures entity-by-entity is time-intensive and inefficient.
Therefore, it is still far from achieving decent procedure text understanding models in both accuracy and efficiency. 

To this end, this paper proposes \textbf{scene-wise} procedural text understanding, a new paradigm that jointly tracks the state changes and locations of all entities scene-by-scene.
Instead of the entity-wise paradigm, we formulate the world described in the procedural text at different timesteps using a sequence of dynamically evolving scenes\footnote{
In procedural text understanding task, the division of timesteps is consistent with the division of sentences.}.
Figure~\ref{fig:introduction} (b) illustrates the whole process of the scene-wise procedural text understanding.
Specifically, each scene contains concepts (e.g., entities, locations or elements from external knowledge) and their relations at current timestep. 
As the narrative develops, the concepts and relations in the scene are dynamically evolved scene-by-scene.
In this way, the state changes and locations of all entities are jointly exploited and then can be simultaneously derived from the scenes. 

Based on this paradigm, we propose \textbf{S}cene \textbf{G}raph \textbf{R}easoner (\textbf{SGR}), a specific implementation for scene-wise procedural text understanding. 
SGR uses a graph structure to model scene. 
Each node in the graph represents a concept, and each edge in the graph represents a relation between two concepts. 
Then the scene evolution is modeled by the graph evolution throughout the whole procedure. 
Specifically, \textbf{SGR} consists of four basic components: 
1) \textbf{a graph structure encoder}, which summarizes critical information from the current scene graph; 
2) \textbf{a context encoder}, which captures the new events occurring from the sentence describing next narrative timestep; 
3) \textbf{a graph structure predictor}, which predicts the evolution of the scene graph after the new events occurring; 
4) \textbf{a state reasoner}, which distills the state changes and locations via comparing the adjacent scene graphs. 
By jointly exploiting all concepts and their relations in the scene graphs, SGR is able to better capture their interactions and associations throughout the whole procedure, and therfore enables to track the state changes and locations of all entities simultaneously in a graph evolution process.

Generally, the main contributions of this paper are:
\begin{itemize}
\item We propose a new scene-wise paradigm for procedural text understanding, which jointly tracks the state changes and locations of all entities scene-by-scene.
\item We design a specific implementation SGR for scene-wise procedural text understanding, which can fully consider the interactions of multiple entities, as well as the associations of state changes and locations. 
\item We conduct experiments on ProPara~\cite{Mishra-2018-Tracking} and Recipes~\cite{Bosselut-2018-Simulating}, two of the representative procedural text understanding benchmarks.
Experiments show that SGR in the scene-wise paradigm achieves the new state-of-the-art procedural text understanding performance, and the reasoning speed is significantly accelerated. 
\end{itemize}

\begin{figure*}[!t]
\centering
\includegraphics[width=0.95\textwidth]{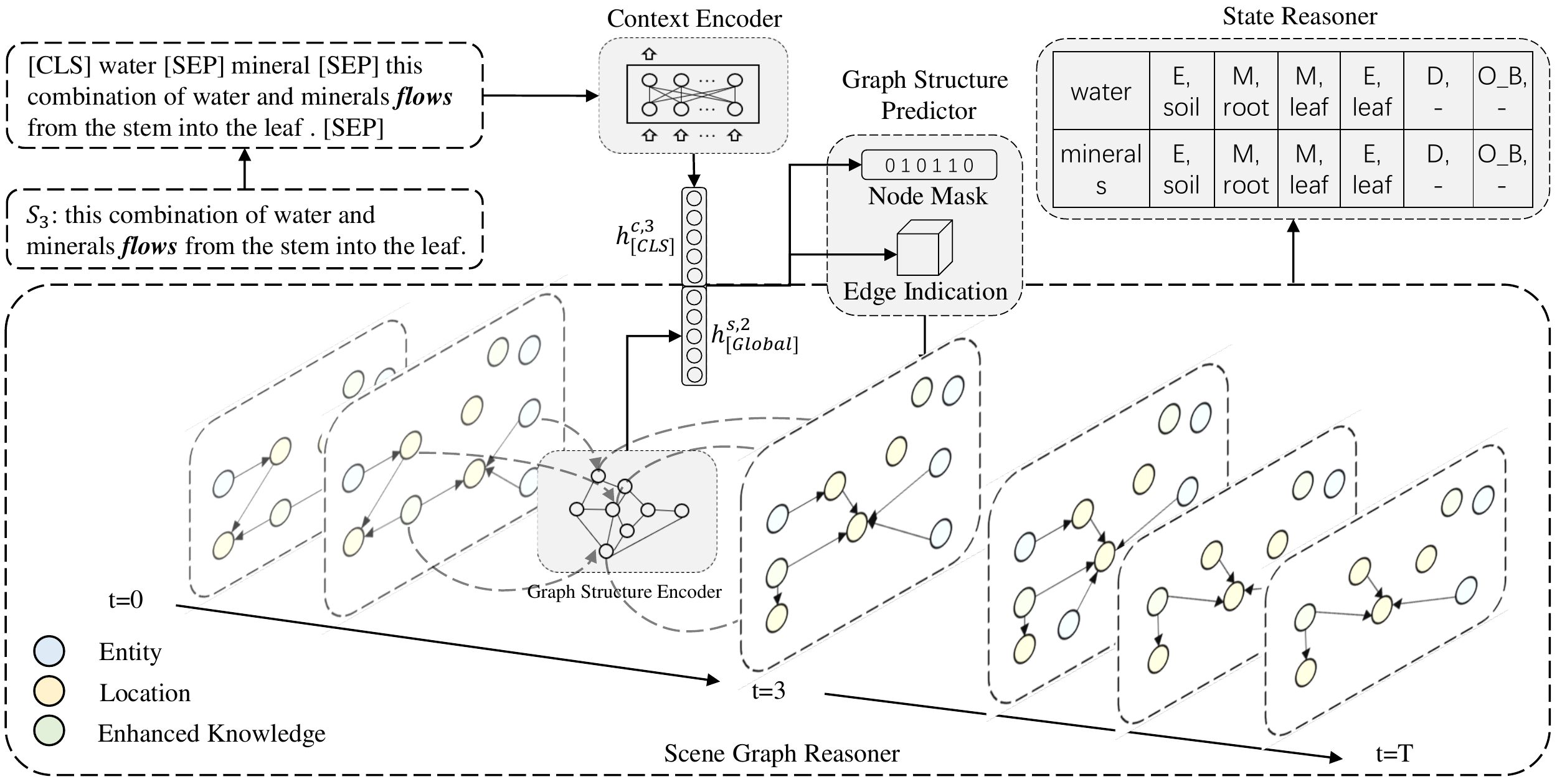}
\caption{An overview of the proposed SGR in the scene-wise paradigm, which is composed of four parts: (a) graph structure encoder; (b) context encoder; (c) graph structure predictor and (d) state reasoner.}
\label{fig:overview}
\end{figure*}

\section{Backgrounds}

\subsection{Task Definition} 
In this paper, we focus on ProPara~\cite{Mishra-2018-Tracking}, which includes a variety of natural procedures, and the task is to answer the questions about the state changes and locations of the entities.
Specifically, given:
\begin{itemize}
\item A paragraph $P$ consists of $T$ sentences $\{S_1, S_2,..., S_T\}$;
\item A set of pre-specified entities $E = \{e_1, e_2,..., e_N\}$ need to be tracked;
\end{itemize}
the procedural text understanding model is required to reason with the described world, and output:
\begin{itemize}
\item State change sequences $Y^s = \{Y_{e_1}^s, Y_{e_2}^s, ..., Y_{e_N}^s\}$ for all pre-specified entities $E$, where $Y_{e_i}^s = \{y_{e_i,1}^s, y_{e_i,2}^s, ..., y_{e_i,T}^s\}$, $y_{e_i,t}^s \in $ \{\emph{Other (O)}, \emph{Exist (E)}, \emph{Move (M)}, \emph{Create (C)}, \emph{Destroy (D)}\}\footnote{\emph{Other (O)} is further devided into \emph{$O_A$, $O_B$}, which mean none state before and after existence separately.}. 
\item Location sequences $Y^l = \{Y_{e_1}^l, Y_{e_2}^l, ..., Y_{e_N}^l\}$ for all pre-specified entities $E$, where $Y_{e_i}^l = \{y_{e_i,1}^l, y_{e_i,2}^l, ..., y_{e_i,T}^l\}$, $y_{e_i,t}^l$ is a text span in the paragraph. A special ``\emph{?}'' token indicates the location is unknown.
\end{itemize}

\subsection{Entity Recognition and \\Location Candidates Generation}
In procedural text understanding, identifying entities is necessary because they are participants in the narrative.
Thus, we first use SpaCy to tokenize the paragraph and all entities.
All text are cleaned and lower-cased.
And then the simple string matching algorithm is used to recognize entities.

Unlike entities, location information in this task is not given initially.
Due to the difficulty to consider arbitrary text spans as possible locations, we follow the previous works~\cite{Gupta-2019-Tracking,Zhang-2020-Knowledge} to generate candidates, and transform the original text span extraction into the candidate classification for tracking locations.
Specifically, we first extract the POS tags by flair~\cite{Akbik-2019-Flair}, and then generate location candidates by POS-based rules\footnote{See details at https://github.com/ytyz1307zzh/NCET-ProPara.}.

For the train and dev sets, if the gold location is not included in the candidates, we manually add them to the candidate set. 
This is mainly for expanding the size of trainable instances in location prediction. 
For the test set, we do not use such method because we obviously cannot know the gold location while testing.

\section{Scene Graph Reasoner}
In this section, we describe how to train an effective procedural text understanding model in the scene-wise paradigm, and track state changes and locations of all entities simultaneously.
As illustrated in Figure~\ref{fig:overview}, we propose \textbf{S}cene \textbf{G}raph \textbf{R}easoner (\textbf{SGR}), a specific implementation for scene-wise procedural text understanding. 
SGR constructs scene graphs for each training instance.
To evolve the scene graphs, SGR first summarizes critical information from the current scene graph by a graph structure encoder, captures the new events occurring from the sentence describing next narrative timestep by a context encoder, and then predicts the evolution of the scene graph after the new events by a graph structure predictor.
During testing, SGR utlizes a state reasoner to simultaneously distills the state changes and locations of all entities via comparing the adjacent scene graphs. 

\subsection{Scene Graph Construction for Training}
For each training instance, we transform the original gold state change and location annotations $\{Y^s, Y^l\}$ into a sequence of scene graphs $Y^g$ to adapt for the proposed scene-wise paradigm, where $Y^g = \{y_1^g, y_2^g, ..., y_t^g, ..., y_T^g\}$.
Each node in the scene graph represents a concept (entities, locations or elements from external knowledge), and each edge represents a predefined relation between two concepts.
As the narrative develops, the nodes and edges in the scene graphs are dynamically created or deleted.
Enlightened by~\cite{Skardinga-2021-Foundations}, we utilize a complete graph with two matrices to record the dynamics of nodes and edges.
And each scene graph can be represented as $y_t^g = \{\hat{\mathcal{G}}, Mask^t, Rel^t\}$: $\hat{\mathcal{G}}$ for the complete graph, $Mask^t$ for node masking and $Rel^t$ for edge indicating at time $t$.
Consequently, the model training objectives is reformulated to predict these scene graphs.

Specifically, SGR first uses the recognized entities and the generated location candidates as nodes\footnote{
It is worth to notice that we do not treat events/actions as one kind of nodes as~\citet{Huang-2021-Reasoning} dose, because events are more suitable for evolving scene graphs.}, 
and use three SRL-based relations (\emph{entity-entity, location-location and entity-location}) as edges to construct the complete graph.
Then SGR enhances the complete graph with the external commonsense knowledge from ConceptNet~\cite{Speer-2017-Conceptnet} because it can provide abundant concept relation, and help the model to understand composition of the world\footnote{
We retrieve and add the corresponding entities and relations (e.g, \emph{HasA}, \emph{PartOf}, and so on) into complete graphs using the same heuristic rules as~\citet{Zhang-2020-Knowledge}.}.
Finally SGR generates the node mask $Mask^t$ and the edge indication $Rel^t$ for each scene graph $y_t^g$, where $Mask^t \in \mathbb{R}^M$ masks the entities that are not created or are already destroied, $Rel^t \in \mathbb{R}^{M*M*R}$ indicates different relations whose arguments are not masked, 
$M$ is the number of concepts, and $R$ is the number of relations.

In this way, the state changes and locations can be jointly modeled in the scene graphs, e.g., the state changes like \emph{Exist}, \emph{Create} and \emph{Destroy} are record by $Mask^t$; the location of each entity are record by $Rel^t$.

\subsection{Graph Structure Encoder for Summarizing Scenes} 
At timestep $t$, SGR adopts a graph attention network (GAT)~\cite{Velickovic-2018-Graph} to summerize critical information from the current scene graph $y_t^g$ since its strong representation capacity. 
In this way, the entities and their states/locations are jointly modeled by rich types of relations among different concepts.

Specifically, the input to the graph attention network is a set of node features $H = \{\hat{h}_1^{s,t}, \hat{h}_2^{s,t}, ..., \hat{h}_M^{s,t}\}$ at timestep $t$, where $M$ is the number of concepts\footnote{
At timestep 0, node features are initialized by context encoder.}.
SGR then performs self-attention on the nodes --- a shared masked attentional mechanism computes attention coefficients:
\begin{equation}
e_{ij}^{t} = a(\mathbf{W_1} \hat{h}_i^{s, t}, \mathbf{W_1} \hat{h}_j^{s, t}, \mathbf{W_2} Rel_{ij}^{t})
\end{equation}
that indicate the importance of node $j$ to node $i$, where $Rel_{ij}^{t}$ is the relation embedding, $\mathbf{W_1}, \mathbf{W_2}$ are learnable parameters.
To make coefficients easily comparable across different nodes, we normalize them across all choices of $j$ using the softmax function:
\begin{equation}
\alpha_{ij}^{t} = softmax_j(e_{ij}^{t}) = \frac{exp(e_{ij}^{t})}{\sum_{k \in \mathcal{N}_i^{t}} exp(e_{ik}^{t})}
\end{equation}
where $\mathcal{N}_i^{t}$ is the neighborhood of node $i$ in the scene graph at timestep $t$, which is determined by the complete graph $\hat{\mathcal{G}}$, the node mask $Mask^t$ and the edge indication $Rel^t$.
Once obtained, the normalized attention coefficients are used to compute a linear combination of the features corresponding to them, to serve as the final features for every node at timestep $t$:
\begin{equation}
h_i^{s, t} = \sigma(\sum_{j \in \mathcal{N}_i^{t}} \alpha_{ij}^{t} \hat{h}_j^{s, t})
\end{equation}

Finally, we obtain the hidden state corresponding to the special $[Global]$ node as the graph structure representation:
\begin{equation}\label{equ:structure}
h^{s,t}_{[Gloabl]} = GAT(y_{t}^g) = GAT(\{\hat{\mathcal{G}}, Mask^t, Rel^t\})
\end{equation}
where $h^{s,t}_{[Gloabl]}$ summerizes critical information from the current scene graph before new events occur.
The scene graph structure and the enhanced external knowledge can be fully learned by the graph structure encoder.

\subsection{Context Encoder for Capturing New Events}
Existing procedural text understanding models do have the context encoder to obtain document-entity representations~\cite{Mishra-2018-Tracking,Mishra-2019-Everything,Du-2019-Consistent,Tang-2020-Understanding,Gupta-2019-Tracking}.
However, we leverage the power of context encoder differently.
We utilize it to capture the new events occurring in the sentence at the next timestep $t+1$.
In this paper, we use BERT~\cite{Devlin-2019-Bert} to handle the nuances of procedural texts.
As suggested by~\citet{Gupta-2019-Effective}, we restructure the input to guide the transformer model~\cite{Vaswani-2017-Attention} to focus on particular entities mentioned in the sentence.

Specifically, take $S_3$ in Figure~\ref{fig:overview} as an example, we first restructure the input as: 
\{\emph{[CLS] water [SEP] minerals [SEP] This combination of water and minerals flows from the stem into the leaf . [SEP]}\}, where \emph{[CLS]} and \emph{[SEP]} are special tokens.
In this way, the transformer can always observe the entities it should be primarily ``attending to'' from the standpoint of building representations. 
For each token in the input, its representation is constructed by concatenating the corresponding token and position embeddings.
Then, the context representation will be inputted into BERT architecture~\cite{Devlin-2019-Bert}, and updated by multilayer Transformer blocks~\cite{Vaswani-2017-Attention}.

Finally, we obtain the hidden state corresponding to the special \emph{[CLS]} token in the last layer as the context representation:
\begin{equation}\label{equ:context}
h^{c,t+1}_{[CLS]} = BERT(restructure(S_{t+1}))
\end{equation}
where $h^{c,t+1}_{[CLS]}$ captures the new events occured in the sentence $S_{t+1}$ at the current timestep $t+1$.
The self-attention mechanism of BERT supports the interactions of the multiple relationships between entities, locations and events mentioned in the sentence $S_{t+1}$.
And it can take advantage of the knowledge learned via pretraining.

\subsection{Graph Structure Predictor for Evolving Scenes}
Based on the the structure representation $h^{s,t}_{[Gloabl]}$ and the context representation $h^{c,t+1}_{[CLS]}$, we can predict the new scene graph structure at timestep $t+1$.

Specifically, we generate the new scene graph $y_{t+1}^g$ via predicting the node mask $Mask^{t+1}$ and the edge indication $Rel^{t+1}$ with the guidance of the aggregate representation:
\begin{equation}\label{equ:predict}
\begin{aligned}
\hat{Mask}^{t+1} = f_1(h^{s,t}_{[Gloabl]}, h^{c,t+1}_{[CLS]}) \\
\hat{Rel}^{t+1} = f_2(h^{s,t}_{[Gloabl]}, h^{c,t+1}_{[CLS]})
\end{aligned}
\end{equation}
where $\hat{Mask}^{t+1} \in \mathbb{R}^M$, $\hat{Rel}^{t+1}  \in \mathbb{R}^{M*M*R}$, $f_1, f_2$ are two nonlinear output layers.
And a sequence of scene graphs can be generated through an autoregressive behavior:
\begin{equation}
y_{t+1}^g = SGR(y_{t}^g, S_{t+1})
\end{equation}

\subsection{Model Training}
Given a training corpus with constructed scene graphs (see Section \textbf{Scene Graph Construction for Training.}), SGR can be supervisedly learned by maximum log-likelihood estimation (MLE):
\begin{equation}
\begin{aligned}
\mathcal{L} &= \sum_{t=1}^{T}\sum_{i=1}^{M} Mask_i^t \log{\hat{Mask}_i^t} \\
&+ \sum_{t=1}^{T}\sum_{i=1}^{M}\sum_{j=1}^{M}\sum_{k=1}^{R} Rel_{ijk}^t \log{\hat{Rel}_{ijk}^t} 
\end{aligned}
\end{equation}
where $T$ is the number of timesteps, $M$ is the number of concepts, including entities, location candidates and elements from external knowledge and $R$ is the number of relations.

\subsection{State Reasoner for Tracking All Entities}
\makeatletter
\newenvironment{breakablealgorithm}
  {
   \begin{center}
     \refstepcounter{algorithm}
     \hrule height.8pt depth0pt \kern2pt
     \renewcommand{\caption}[2][\relax]{
       {\raggedright\textbf{\ALG@name~\thealgorithm} ##2\par}%
       \ifx\relax##1\relax 
         \addcontentsline{loa}{algorithm}{\protect\numberline{\thealgorithm}##2}%
       \else 
         \addcontentsline{loa}{algorithm}{\protect\numberline{\thealgorithm}##1}%

       \fi
       \kern2pt\hrule\kern2pt
     }
  }{
     \kern2pt\hrule\relax
   \end{center}
  }
\makeatother
{
\renewcommand
\baselinestretch{1.1}
\begin{algorithm}[!t]
\footnotesize
\renewcommand{\algorithmicrequire}{\textbf{Input:}}
\renewcommand\algorithmicensure {\textbf{Return:} }
\caption{: State Reasoner.}
\label{alg:reasoner}
\begin{algorithmic}[1]
\REQUIRE
SGR: the trained procedural text understanding model;\\
ConceptNet: the external commonsense knowledge base;\\
$P = \{S_1, S_2,..., S_T\}$: the procedural text;\\
$E = \{e_1, e_2,..., e_N\}$: pre-specified entities;\\
$Constraints$: used for postprocessing;
\STATE The complete graph $\mathcal{G}$ $\leftarrow$ \textbf{Construct}($P, E$)
\STATE The enhanced complete graph $\hat{\mathcal{G}}$ $\leftarrow$ \textbf{Enhance}($\mathcal{G}$, ConceptNet)
\STATE $Y^g$ $\leftarrow$ $\emptyset$
\STATE $y_0^g$ $\leftarrow$ ($\hat{\mathcal{G}}$, $Mask^0$, $Rel^0$) $\leftarrow$ ($\hat{\mathcal{G}}$, $\emptyset$, $\emptyset$)
\STATE $Y^g\textbf{.append}(y_0^g)$
\STATE \textbf{for} $S_t$ in $P$ \textbf{do}
\STATE \ \ \ \ \ \ $h^{s,t}_{[Gloabl]}$ $\leftarrow$ $\textbf{Graph Structure Encoder}(y_{t}^g)$
\STATE \ \ \ \ \ \ $h^{c,t+1}_{[CLS]}$ $\leftarrow$ $\textbf{Context Encoder}(S_{t+1})$ 
\STATE \ \ \ \ \ \ $y_{t+1}^g$ $\leftarrow$ $(\hat{\mathcal{G}}, \hat{Mask}^t, \hat{Rel}^t)$ $\leftarrow$ \\ 
\ \ \ \ \ \ \ \ \ \ \ \ $\textbf{Graph Structure Predictor}(h^{s,t}_{[Gloabl]}, h^{c,t+1}_{[CLS]})$
\STATE \ \ \ \ \ \ $Y^g\textbf{.append}(y_{t+1}^g)$
\STATE $(Y^s, Y^l)$ $\leftarrow$ $\textbf{Transform}(\hat{\mathcal{G}}, Y^g, Constraints)$
\ENSURE $Y^s, Y^l$;
\end{algorithmic}
\end{algorithm}
\par}

During testing, on the basis of the trained procedural text understanding model SGR, we can construct scene graphs for new narratives, and then simultaneously track the state changes and locations of all entities scene-by-scene.
Specifically, we can infer the state change and location sequences via comparing the adjacent scene graphs, e.g., if we find out that ``\emph{water}'' is not masked at scene $y_{t-1}^g$ but masked at scene $y_t^g$, we can deterministically infer the state of ``\emph{water}'' is ``\emph{Destroy (D)}'' at timestep $t$; if we find that ``\emph{water}'' has a ``\emph{LocateIn}'' realtion with ``\emph{root}'', the current location of ``\emph{water}'' must be ``\emph{root}''.
It is worth to notice that these transformations of all entities can be processed in parallel.

To facilitate the description of state reasoner, we summarize this process in Algorithm~\ref{alg:reasoner}.
Specifically, we first preprocess the raw input $\{P, E\}$ to construct the complete graph, and then utilize the external commonsense knowledge ConceptNet~\cite{Speer-2017-Conceptnet} to get the enriched complete graph (\textbf{Line 1-2}).
Second, we initialize the scene graphs $Y^g$ as $\emptyset$ (\textbf{Line 3}).
However, during testing, we cannot access the gold state change and location annotations.
Thus, we initialize $Mask^0$ and $Rel^0$ as zeros, which means that we know nothing about the current world at timestep $0$ (\textbf{Line 4}).
After these preparations, we utilize the trained model SGR to evolve the scene graphs from $y_0^g$ to $y_T^g$, which contains graph structure encoding, context encoding and graph structure predicting (\textbf{Line 6-10}).
Finally, we transform the scene graphs $Y^g$ into the state change sequence $Y^s$ and the location sequence $Y^l$ for all pre-specified entities (\textbf{Line 6-11}).
The constraints used in previous works can be easily inject into the final transformation process, e.g., correct invalid actions according to the whole action sequence~\cite{Tang-2020-Understanding}.

In this way, the state change and location sequences of all entities can be tracked simultaneously, and the efficiency of reasoning can be significantly improved.

\section{Experiments}
\begin{table}[!t]
\centering
\resizebox{0.95\columnwidth}{!}{
\begin{tabular}{l|ccc|ccc}  
\hline 
\multirow{2}*{\textbf{Statistics}} & \multicolumn{3}{c|}{\textbf{ProPara}} & \multicolumn{3}{c}{\textbf{Recipes}} \\
\cline{2-7}
 & \textbf{train} & \textbf{dev} & \textbf{test} & \textbf{train} & \textbf{dev} & \textbf{test} \\
\hline 
\hline 
\#Instance/Para & 391 & 43 & 54 & 693 & 86 & 87 \\
\#Sentence & 2,620 & 288 & 372 & 6,098 & 765 & 783 \\
\#Entity & 1,504 & 175 & 236 & 5,932 & 756 & 737 \\
\hline 
Avg.\#sent/para & 6.7 & 6.7 & 6.9 & 8.8 & 8.9 & 9.0 \\
Avg.\#enti/para & 3.8 & 4.1 & 4.4 & 8.6 & 8.8 & 8.5 \\
\hline 
\end{tabular}
}
\caption{Statistics of ProPara and Recipes datasets.
We regard one paragraph with all pre-specified entities as an instance.
Thus, the number of instances is equivalent to the number of paragraphs.}
\label{tab:statistics}
\end{table}

\begin{table*}[!t]
\centering
\resizebox{0.95\textwidth}{!}{
\begin{tabular}{l|ccc|ccccc}  
\hline 
\multirow{2}*{\textbf{Models}} & \multicolumn{3}{c|}{\textbf{Document-level task}} & \multicolumn{5}{c}{\textbf{Sentence-level task}} \\
 & Precision & Recall & F1 & Cat-1 & Cat-2 & Cat-3 & Macro-Avg & Micro-Avg\\
\hline
\hline 
\textbf{Entity-wise Models} & & & & & & & & \\
\quad \textbf{Models with context encoder} & & & & & & & & \\
\quad \quad \ EntNet~\cite{Henaff-2017-Tracking} & 54.7 & 30.7 & 39.4 & 51.6 & 18.8 & 7.8 & 26.1 & 26.0 \\
\quad \quad \ QRN~\cite{Seo-2017-Bidirectional} & 60.9 & 31.1 & 41.4 & 52.4 & 15.5 & 10.9 & 26.3 & 26.5 \\
\quad \quad \ ProLocal~\cite{Mishra-2018-Tracking} & 81.7 & 36.8 & 50.7 & 62.7 & 30.5 & 10.4 & 34.5 & 34.0 \\
\quad \quad \ ProGlobal~\cite{Mishra-2018-Tracking} & 48.8 & 61.7 & 51.9 & 63.0 & 36.4 & 35.9 & 45.1 & 45.4 \\
\quad \ \ $\bullet$ AQA~\cite{Ribeiro-2019-Predicting} & 62.0 & 45.1 & 52.3 & 61.6 & 40.1 & 18.6 & 39.4 & 40.1 \\
\quad \ \ $\bullet$ ProStruct~\cite{Tandon-2018-Reasoning} & 74.3 & 43.0 & 54.5 & - & - & - & - & - \\
\quad \quad \ XPAD~\cite{Mishra-2019-Everything} & 70.5 & 45.3 & 55.2 & - & - & - & - & - \\
\quad \quad \ LACE~\cite{Du-2019-Consistent} & 75.3 & 45.4 & 56.6 & - & - & - & - & - \\
\quad \quad \ NCET~\cite{Gupta-2019-Tracking} & 67.1 & 58.5 & 62.5 & 73.7 & 47.1 & 41.0 & 53.9 & 54.0 \\
\quad \ \ $\diamond$ ET\_BERT~\cite{Gupta-2019-Effective} & - & - & - & 73.6 & 52.6 & - & - & - \\
\quad \ \ $\ast$ IEN~\cite{Tang-2020-Understanding} & 69.8 & 56.3 & 62.3 & 71.8 & 47.6 & 40.5 & 53.3 & 53.0 \\
\quad \ \ $\diamond$ DYNAPRO~\cite{Amini-2020-Procedural} & 75.2 & 58.0 & 65.5 & 72.4 & 49.3 & \underline{\textbf{44.5}} & 55.4 & 55.5 \\
\ \ \ $\bullet$ $\diamond$ KOALA~\cite{Zhang-2020-Knowledge} & 77.7 & \underline{\textbf{64.4}} & 70.4 & 78.5 & 53.3 & 41.3 & 57.7 & 57.5 \\
\quad \textbf{Models with structure encoder} & & & & & & & & \\
\quad \quad \ KG-MRC~\cite{Das-2019-Building} & 69.3 & 49.3 & 57.6 & 62.9 & 40.0 & 38.2 & 47.0 & 46.6 \\
\ \ \quad $\bullet$ ProGraph~\cite{Zhong-2020-Heterogeneous} & 67.3 & 55.8 & 61.0 & 67.8 & 44.6 & 41.8 & 51.4 & 51.5 \\
\ \ \quad $\diamond$ TSLM~\cite{Faghihi-2021-Time} & 68.4 & 68.9 & 68.6 & 78.8 & 56.8 & 40.9 & 58.8 & 58.3 \\
\ \ \quad $\diamond$ REAL~\cite{Huang-2021-Reasoning} & 81.9 & 61.9 & 70.5 & 78.4 & 53.7 & 42.4 & 58.2 & 57.9 \\
\hline 
\textbf{Scene-wise Models} & & & & & & & & \\
$\bullet$ $\diamond$ $\ast$ SGR (our method) & \underline{\textbf{84.9}} & 62.9 & \underline{\textbf{72.2}} & \underline{\textbf{79.9}} & 55.1 & 43.5 & \underline{\textbf{59.5}} & \underline{\textbf{59.2}} \\
\quad \quad \quad w/o Graph Structure Encoder & 72.4 & 51.1 & 59.9 & 69.9 & 42.7 & 39.9 & 50.8 & 51.0 \\
\quad \quad \quad w/o Context Encoder  & 76.1 & 55.4 & 64.1 & 74.9 & 47.9 & 40.0 & 54.3 & 54.2 \\
\quad \quad \quad w/o ConceptNet & 82.7 & 63.2 & 71.6 & 78.3 & \underline{\textbf{56.0}} & 42.5 & 58.9 & 58.6 \\
\quad \quad \quad w/o Pre-trained Bert & 81.8 & 59.2 & 68.7 & 76.2 & 53.3 & 41.4 & 57.0 & 56.7 \\
\hline 
\end{tabular}
}
\caption{Experimental results on ProPara document-and sentence-level tasks.
$\ast$, $\bullet$ and $\diamond$ indicate the models consider the interactions between multiple entities, use the external knowledge base and are equipped with the pre-trained model separately.}
\label{tab:ProPara_results}
\end{table*}

\begin{table}[!t]
\centering
\resizebox{0.95\columnwidth}{!}{
\begin{tabular}{l|ccc}  
\hline 
\textbf{Models} & \textbf{Precision} & \textbf{Recall} & \textbf{F1} \\
\hline 
\hline 
NCET (re-implementation) & 56.5 & 46.4 & 50.9 \\
IEN (re-implementation) & 58.5 & 47.0 & 52.2 \\
KOALA~\cite{Zhang-2020-Knowledge} & 60.1 & 52.6 & 56.1 \\
REAL~\cite{Huang-2021-Reasoning} & 55.2 & \underline{\textbf{52.9}} & 54.1 \\
\hline 
SGR (our method) & \underline{\textbf{69.3}} & 50.5 & \underline{\textbf{58.4}} \\
\hline 
\end{tabular}
}
\caption{Experimental results on re-split Recipes.}
\label{tab:recipes_results}
\end{table}

\subsection{Experimental Settings}
\textbf{Dataset.}
We conduct main experiments on ProPara~\cite{Mishra-2018-Tracking} and auxiliary experiments on Recipes~\cite{Bosselut-2018-Simulating}.
For ProPara, we follow the official split~\cite{Mishra-2018-Tracking} for train/dev/test set. 
For Recipes, following the previous works~\cite{Zhang-2020-Knowledge,Huang-2021-Reasoning}, we only use the human-labeled data in our experiments, and re-split it into 80\%/10\%/10\% for train/dev/test sets. 
More statistics about these two datasets are shown in Table~\ref{tab:statistics}\footnote{
The number of instances is different from the previous entity-wise works because they regard one entity-paragraph pair as an instance, and result in 1.9k instances}. 

\textbf{Implementation Details.}
For graph structure encoder, we apply a one-layer graph attention network (GAT)~\cite{Velickovic-2018-Graph}.
For context encoder, we use the BERT base implemented by HuggingFace's transformers library~\cite{Wolf-2020-Transformers}. 
Hyper-parameters are manually tuned according to the accuracy on the dev set: 
batch size is set to 16, hidden size is set to 128 and learning rate is set to 5e-5.
The final model is trained on an Nvidia TITAN RTX GPU with Adam optimizer~\cite{Kingma-2015-Adam}, and is selected with the highest prediction accuracy on dev set.

\textbf{Evaluation Metrics}
For ProPara, following the previous works, we perform document level~\cite{Tandon-2018-Reasoning} and sentence level~\cite{Mishra-2018-Tracking} tasks in our main experiments.
Specifically, the document level task requires models to answer the four document-level questions:
\begin{description}
\item[\textbf{Q1}:] What are the inputs to the procedure? 
\item[\textbf{Q2}:] What are the outputs of the procedure? 
\item[\textbf{Q3}:] What conversions occur, when and where? 
\item[\textbf{Q4}:] What movements occur, when and where? 
\end{description}
The evaluator compute precision, recall and F1 score for each question, and the overall F1 score is the macro-average of the above four questions\footnote{\url{https://github.com/allenai/aristo-leaderboard/tree/master/ProPara}}.
The sentence-level task requires models to answer ten fine grained sentence-level questions, which can be summarized into three categories:
\begin{description}
\item[\textbf{Cat-1}:] Is entity created (destroyed, moved)? 
\item[\textbf{Cat-2}:] When is entity created (destroyed, moved)?
\item[\textbf{Cat-3}:] Where is entity created (destroyed, moved from/to)? 
\end{description}
Evaluation metrics are macro-average and micro-average accuracy of three sets of questions.
More details can be found in the official script\footnote{\url{https://github.com/allenai/ProPara/tree/master/ProPara/evaluation}}.
The answers of both document- and sentence-level questions can be deterministically computed from the state change and location sequences.

For Recipes, we follow~\citet{Zhang-2020-Knowledge,Huang-2021-Reasoning} to predict the location changes of the ingredients during the procedure. 
For each movement, the model should predict the new location of the entity, plus the timestep when the movement occurs. 
We take precision, recall, and F1 scores to evaluate models.

\subsection{Baselines}
For ProPara, we compare SGR with the following baselines, most of them are on the official leaderboard\footnote{\url{https://leaderboard.allenai.org/ProPara/submissions/public}}:
\begin{itemize}
\item \textbf{Models with context encoder} rely on Bi-LSTM/Bert to obtain the document-entity representations and track the states/locations separately.
\item \textbf{Models with structure encoder} leverage the power of the static graph to obtaion more effective document representations. Different from our work, they lack the dynamical representations of the procedures.
\end{itemize}
For Recipes, we compare SGR with the state-of-the-art models: NCET~\cite{Tang-2020-Understanding}, KOALA~\cite{Zhang-2020-Knowledge} and REAL~\cite{Huang-2021-Reasoning}.

\subsection{Overall Results}
Table~\ref{tab:ProPara_results},~\ref{tab:recipes_results} and~\ref{tab:speed_results} show the overall results.
We can see that: 

\textbf{1. The proposed SGR in the scene-wise paradigm achieves the state-of-the-art performance.}
SGR can significantly outperform the state-of-the-art model REAL and achieves 72.2 F1 on ProPara document-level task and 59.5/59.2 Macro-Avg/Micro-Avg scores on ProPara sentence-level task. 
On Recipes, SGR also outperforms the corresponding baselines and achieves 58.4 F1.
We believe this is because that the entities and their states/locations are jointly modeled in the scenes.
Therefore, the association of two track targets and the interaction of the multiple entities are fully explored.

\textbf{2. Reasoning states and locations of all entities scene-by-scene significantly improves the inference efficiency.}
We compare the inferencing time of SGR with NCET and IEN on an Nvidia TITAN RTX GPU in Table~\ref{tab:speed_results}. 
NCET is the traditional model in the entity-wise paradigm, while IEN considers the interactions between multiple entities via the entity-location attention mechanism\footnote{
Other state-of-the-art models spend more reasoning times than NCET and IEN due to exquisitely designed archectures.}.
However, both NCET and IEN rely on separate trackers with CRF to predict state changes and locations.
Thus they are time-intensive and take almost 3-4 times as long as SGR to track each entity.
These results verifies that reasoning questions scene-by-scene is efficient and promotes the application of the procedural text understanding model in real-world scenes.

\textbf{3. The graph structure encoder and the context encoder are indispensable, and are complementary with each other.}
When compared with the full model SGR, its two variants SGR w/o Graph Structure Encoder and SGR w/o Context Encoder show declined performance in different degrees, which indicates that the current scene modeled by the graph structure encoder and the new events captured by the context encoder are necessary. 
Surprisingly, we find that SGR w/o Context Encoder still perform quite well.
The insight in those observations may be that it can be regard as a graph-structured language model --- predicts snapshots through an autoregressive behavior, and builds label consistency in the same topic~\cite{Du-2019-Consistent}.

\begin{table}[!t]
\centering
\resizebox{0.95\columnwidth}{!}{
\begin{tabular}{l|ccc}  
\hline 
\multirow{2}*{\textbf{Models}} & \multicolumn{3}{c}{\textbf{ProPara}}\\
 & Total & Avg./para & Avg./enti \\
\hline 
\hline 
NCET (re-implementation) & 51.31 & 0.95 & 0.22 \\
IEN (re-implementation) & 42.50 & 0.79 & 0.18 \\
\hline 
SGR (our method) & \underline{\textbf{17.99}} & \underline{\textbf{0.33}} & \underline{\textbf{0.08}} \\
\hline 
\end{tabular}
}
\caption{Inference time (seconds) on ProPara.}
\label{tab:speed_results}
\end{table}

\begin{table}[!t]
\centering
\resizebox{0.95\columnwidth}{!}{
\begin{tabular}{l|cc|ccc}  
\hline 
& \multicolumn{2}{c|}{\textbf{ConceptNet}} & \multicolumn{3}{c}{\textbf{Document-level task}}\\
& Train & Test & Precision & Recall & F1 \\
\hline 
\hline 
SGR $\vartriangle$ & \checkmark & \checkmark  & \underline{\textbf{84.9}} & 62.9 & \underline{\textbf{72.2}} \\
SGR(test) $\vartriangle$ & & \checkmark & 83.7 & 62.9 & 71.8 \\
SGR(train) $\triangledown$ & \checkmark & & 81.5 & 60.4 & 69.4 \\
SGR(none) & & & 82.7 & \underline{\textbf{63.2}} & 71.6 \\
\hline 
\end{tabular}
}
\caption{Effect of the usage of ConceptNet on ProPara.
All improvements of SGR are statistical significance at p\textless0.01.}
\label{tab:detail_results}
\end{table}

\subsection{Detailed Analysis}
The external knowledge can be easily injected into SGR.
To investigate the effectiveness of difference kinds of external knowledge, we design the following experiments.

\textbf{Effects of the External Knowledge from the Pre-trained language model.}
From Table~\ref{tab:ProPara_results}, we can see that models indicates with $\diamond$ outperform other baselines.
When compared with SGR, the performance of SGR w/o Pre-trained BERT clampes between SGR and SGR w/o Context Encoder.
It means that not only the context encoder but also the external knowledge learned via pre-training is helpful for procedural text understanding.

\textbf{Effects of the External Knowledge from the Knowledge Base.}
First of all, from Table~\ref{tab:ProPara_results}, we can see that models indicates with $\bullet$ outperform other baselines.
And the decay of SGR w/o ConceptNet is also appreciable when compared with SGR.
These results verify the effectiveness of the external knowledge from knowledge base.
Furthermore, we investigate the usage of ConceptNet in Table~\ref{tab:detail_results}.
We can see that: 
1) Compared with SGR(none), SGR and SGR(test) lead to improvements.
The reason behinds it is that ConceptNet can constrain the prediction space of graph structure predictor and help the model to understand composition of the world.
2) SGR(train) even perform worse than SGR(none).
It is because that the inconsistency between train and test introduces too many noisies rather than knowledge.
In other words, the exposure bias of the autoregressive behavior hurts the performance of the model~\cite{Zhang-2019-Bridging}.

\section{Related Work}

\textbf{Procedural Text Understanding} is important and challenging.
Many datasets have been proposed such as bAbI~\cite{Henaff-2017-Tracking}, RECIPES~\cite{Kiddon-2015-Mise} and ProPara~\cite{Mishra-2018-Tracking}.
Blessed with valuable benchmarks, there emerge abundant procedural text understanding models which are in the question-answering framework~\cite{Henaff-2017-Tracking,Seo-2017-Bidirectional,Das-2019-Building} or hierarchical neural network framework~\cite{Mishra-2018-Tracking,Tandon-2018-Reasoning,Du-2019-Consistent,Gupta-2019-Tracking,Gupta-2019-Effective,Zhang-2020-Knowledge,Zhong-2020-Heterogeneous}.
Some of them utilize graph encoder to obtaion more effective document-entity representations~\cite{Huang-2021-Reasoning} and almost of them in the entity-wise paradigm.
Different from them, this paper propose a new scene-wise paradigm to jointly tracks the state changes and locations of all entities scene-by-scene.

\textbf{Dynamic Graph Neural Networks (DGNNs)} are used in a wide range of fields, including social network analysis, recommender systems and epidemiology~\cite{Yin-2019-Graph,Skardinga-2021-Foundations}.
DGNNs add a new dimension to network modeling and prediction – time. 
This new dimension radically influences network properties which enable a more powerful representation of network, and increases predictive capabilities of methods~\cite{Aggarwal-2014-Evolutionary,Li-2018-Restricted}.
In this paper, we utilize a graph structure to model scene, and an evaluation algorithm is proposed to adapt for the proposed scene-wise paradigm.

\section{Conclusions}
In this paper, we propose a new \textbf{scene-wise} paradigm for procedural text understanding and \textbf{S}cene \textbf{G}raph \textbf{R}easoner (\textbf{SGR}) is designed to jointly model the associations of state changes and locations, as well as the interactions of multiple entities.
In this way, the state changes and locations of all entities are jointly exploited and then can be simultaneously derived from the scene graphs. 
Experiments show that SGR achieves the new state-of-the-art procedural text understanding performance, and the reasoning speed is significantly accelerated. 
For future work, we want to pretrain a graph-structured language model to build label consistency in the same topic~\cite{Du-2019-Consistent} and design new training and reasoning methods to overcome the exposure bias of the autoregressive behavior~\cite{Zhang-2019-Bridging}.

\section{Acknowledgments}
We sincerely thank the reviewers for their insightful comments and valuable suggestions.
This research work is supported by the National Key R\&D Program of China (2020AAA0105200), and the National Natural Science Foundation of China under Grants no. U1936207, 62122077, 62106251 and 61772505, the Beijing Academy of Artificial Intelligence (BAAI) and in part by the Youth Innovation PromotionAssociation CAS(2018141).

\bibliography{aaai22}

\end{document}